\pdfoutput=1
\documentclass[11pt]{article}

\usepackage{acl}

\usepackage{times}
\usepackage{latexsym}

\usepackage[T1]{fontenc}
\usepackage[utf8]{inputenc}

\usepackage{microtype}

\usepackage{subcaption}
\usepackage{url}
\usepackage{hyperref}
\usepackage{graphicx}
\usepackage{graphics}
\usepackage{amsmath}
\usepackage{amsthm}
\usepackage{algorithm}
\usepackage{algorithmic}
\usepackage{microtype}
\usepackage{booktabs} % for professional tables
\usepackage{makecell}
\usepackage{arydshln}   % note the strict order of the packages: booktabs, makecell, arydshln
\usepackage{multirow}
\usepackage{bbm}
\usepackage{balance}
\usepackage{amssymb}
\usepackage{textcomp}
\usepackage{color}
\usepackage{xcolor}
\usepackage{enumitem}

\usepackage[export]{adjustbox}

\definecolor{zred}{RGB}{196, 38, 11}
\definecolor{zblue}{RGB}{41, 52, 190}
\definecolor{zgreen}{RGB}{18, 141, 21}

\definecolor{zptu}{RGB}{18, 141, 21}

\newcommand{\oaxe}{\textsc{OaXE}~}
\newcommand{\noaxe}{{\em ngram}-\textsc{OaXE}~}

\title{\noaxe: Phrase-Based Order-Agnostic Cross Entropy for Non-Autoregressive Machine Translation}

\author{
Cunxiao Du\\Singapore Management University\\\texttt{cnsdunm@gmail.com}\\
\And
Zhaopeng Tu\thanks{~~Zhaopeng Tu is the corresponding author.}\\Tencent AI Lab\\\texttt{zptu@tencent.com}\\
\AND
Longyue Wang\\Tencent AI Lab\\\texttt{vinnylywang@tencent.com}\\
\And
Jing Jiang\\Singapore Management University\\\texttt{jingjiang@smu.edu.sg}
}

\begin{document}

\maketitle

\begin{abstract}
Recently, a new training \oaxe loss~\cite{Du:2021:ICML} has proven effective to ameliorate the effect of multimodality for non-autoregressive translation (NAT), which removes the penalty of word order errors in the standard cross-entropy loss. Starting from the intuition that reordering generally occurs between phrases, we extend \oaxe by only allowing reordering between ngram phrases and still requiring a strict match of word order within the phrases. Extensive experiments on NAT benchmarks across language pairs and data scales demonstrate the effectiveness and universality of our approach. 
%Further analyses show that the proposed \noaxe alleviates the multimodality problem with a better modeling of phrase translation.
Further analyses show that \noaxe indeed improves the translation of ngram phrases, and produces more fluent translation with a better modeling of sentence structure.\footnote{The codes and models are in \url{https://github.com/tencent-ailab/machine-translation/COLING22_ngram-OAXE/}.}
\end{abstract}

\section{Introduction}
\label{intro}

Fully non-autoregressive translation (NAT) has received increasing attention for its efficient decoding by predicting every target token in parallel~\cite{NAT,maskp}.
However, such advantage comes at the cost of sacrificing translation quality due to the {\em multimodality} problem:
there exist many possible translations of the same sentence, while vanilla NAT models may consider them at the same time due to the independent predictions, % of target tokens, 
which leads to multi-modal outputs in the form of token repetitions~\cite{NAT}.

Recent works have incorporated approaches to improving the standard cross-entropy (XE) loss to ameliorate the effect of multimodality.
The motivation for these works is that modeling word order is difficult for NAT, since the model cannot condition on its previous predictions like its autoregressive counterpart.
Starting from this intuition, a thread of research relaxes the word order restriction based on the {\bf monotonic alignment} assumption~\cite{libovicky-helcl-2018-end,axe,imputer}.~\newcite{Du:2021:ICML} take a further step by removing the penalty of word order errors with a novel order-agnostic cross entropy (\textsc{OaXE}) loss, which enables NAT models to handle {\bf word reordering} -- a common source of multimodality problem. Accordingly, \oaxe achieves the best performance among these model variants.

However, \oaxe allows reordering between every two words, which is not always valid in practice. For example, the reordering of the two words ``this afternoon'' is not correct in grammar. The reordering generally occurs between ngram phrases, such as ``I ate pizza'' and ``this afternoon''. Starting from this intuition, we extend \oaxe by constraining the reordering between ngrams and requiring a strict match of word order within each ngram (i.e., {\em ngram}-\textsc{OaXE}).
% In this work, we propose to extend \oaxe with a linguistic prior -- reordering generally occurs between phrases. Specifically, the new loss (namely {\em ngram}-\textsc{OaXE}) only allows reordering between ngram phrases, and requires a strict match of word order within the phrases between target tokens and model predictions.
To this end, we first build the probability distributions of ngrams in the target sentence using the word probabilities produced by NAT models.
Then we find the best ordering of target ngrams to minimize the cross entropy loss. We implement the \noaxe loss in an efficient way, which only adds one more line of code on top of the source code of \textsc{OaXE}.
Accordingly, \noaxe only marginally increases training time (e.g., 3\% more time) over \textsc{OaXE}.

\begin{figure*}[t]
    \centering 
    \includegraphics[width=\textwidth]{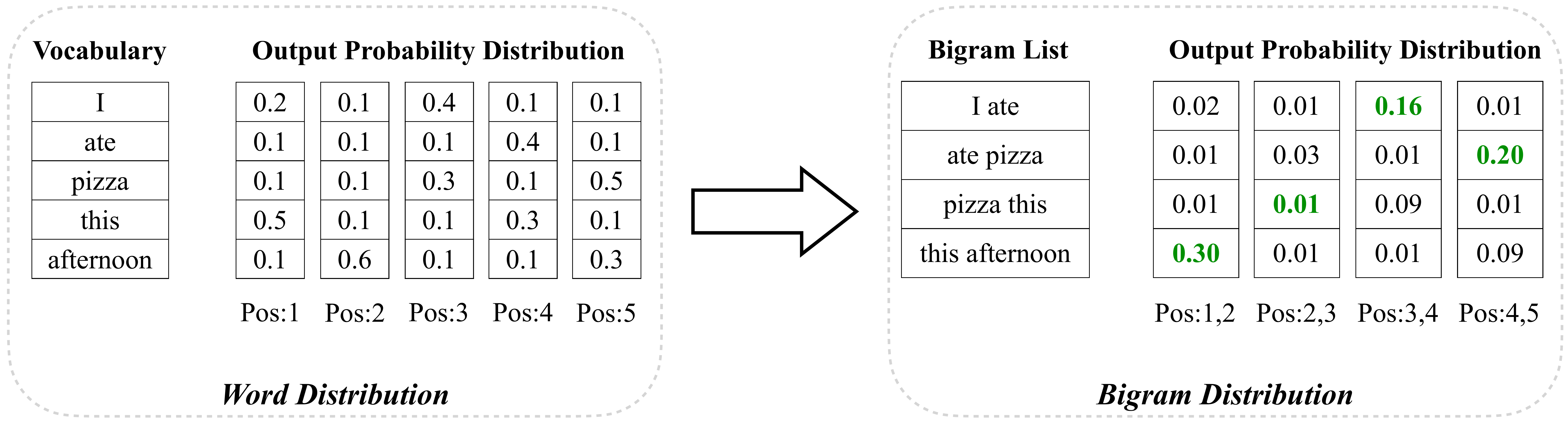} \caption{ Illustration of the proposed \noaxe loss with $N=2$ (i.e., bigram-\textsc{OaXE}). We only show the probabilities of the target words and bigrams for better illustration. Firstly, \noaxe transforms the word probability distributions to the bigram distributions by multiplying the word probabilities at the corresponding positions. For example, P(``I ate" | Pos:1,2) = P(``I'' | Pos:1) * P(``ate'' | Pos:2) = 0.2*0.1=0.02. 
    Then, we select the ngrams (highlighted in bold) for each neighbouring positions using the efficient Hungarian algorithm.}
    \label{fig:ngram-oaxe}
\end{figure*}

Experimental results on widely-used NAT benchmarks show that \noaxe improves translation performance over \oaxe in all cases. % with a better modeling of phrase translation.
Encouragingly, \noaxe outperforms \oaxe by up to +3.8 BLEU points on raw data (without knowledge distillation) for WMT14 En-De translation (Table~\ref{tab:reference}), and narrows the performance gap between training on raw data and on distilled data.
Further analyses show that \noaxe improves over \oaxe on the generation accuracy of ngram phrases and modeling reordering between ngram phrases, which makes \noaxe handle long sentences better, especially on raw data.
The strength of \noaxe on directly learning from the complex raw data indicates the potential to train NAT models without knowledge distillation.

\section{Methodology}

\subsection{Preliminaries: NAT}
\label{sec:preliminary}

\paragraph{Cross Entropy (XE)}
Standard NAT models~\cite{NAT} are trained with the cross entropy loss:
\begin{eqnarray}
    \mathcal{L}_{XE} = -\log P(Y | X)  = - \sum_{y_n} \log P(y_i | X),
    \label{eq:ce}
\end{eqnarray}
where $(X, Y)$ with $Y=\{y_1, \dots, y_I\}$ is a bilingual training example, and $P(y_i | X)$ is calculated independently by the NAT model.
XE requires a strict match of word order between target tokens and model predictions, thus will heavily penalize hypotheses that are semantically equivalent to the target but different in word order.

\paragraph{Order-Agnostic Cross Entropy (\textsc{OaXE})}
~\newcite{Du:2021:ICML} remove the word order restriction of XE, and assign loss based on the best alignment between target tokens and model predictions. 
They define the ordering space ${\bf O}=\{O^1, \dots, O^J\}$ for $Y$, where $O^j$ is an ordering of the set of target tokens $(y_1, \dots, y_I)$.
The \textsc{OaXE} objective is defined as finding the best ordering $O^j$ to minimize the cross entropy loss:
\begin{equation}
    \mathcal{L}_{\textsc{OaXE}} = \underset{O^j \in {\bf O}}{\operatorname{min}} \left(-\log P(O^{j}|X) \right), \label{eq:oace}
\end{equation}
where $-\log P(O^{i}|X)$ is the cross entropy loss for ordering $O^i$, which is calculated by Equation~\ref{eq:ce}.

\subsection{\noaxe Loss}
\label{sec:oace}

Figure~\ref{fig:ngram-oaxe} illustrates the two-phase calculation of \noaxe: 1) constructing the probability distributions of the ngrams in the target sentence; 2) searching the best ordering of the considered ngrams to minimize the cross entropy loss.

\paragraph{Formulation}
Given the target $Y=\{y_1, \dots, y_I\}$, we define the target ngrams $G^N$ of size N as all the $N$ continuous tokens in $Y$: $\{y_{1:N}, \cdots, y_{I-N+1:I}\}$. The output ngram distributions $P_G$ is defined as:
\begin{equation}
P_G(y_{i:i+N-1}|X)=\prod\limits_{t=i}^{i+N-1} P(y_t|X),
\label{eqn:ngram-distributions}
\end{equation}
where $P(y_t|X)$ is the prediction probability of NAT models for the token $y_t$ in position $t$ of the target sentence, and $N$ is the size of ngrams.

The \noaxe objective is defined as finding the best ordering $O^j$ to minimize the cross entropy loss of the considered ngrams in target sentence $Y$:
\begin{eqnarray}
    \mathcal{L}_\text{\noaxe} &= \underset{O^j \in {\bf O}}{\operatorname{min}} \left(-\log {P}_G(O^{j}|X) \right).
    \label{eq:noace}
\end{eqnarray}
Ideally, the best ordering $O^j$ should meet the following conditions:
\begin{itemize}
    \item [1.] The ngrams in $O^j$ should not be overlapped (e.g., ``I ate" and ``ate pizza" should not occur simultaneously in one $O$).
    \item [2.] $O^j$ is a mixture of ngrams with different sizes (e.g., ``I ate pizza" and ``this afternoon").
\end{itemize}
However, it is computationally infeasible to search the best ngram segmentation of the target sentence with highest probabilities. Given a target sentence with length I, there are $2^I$ ngram segmentation (i.e, each token can be labeled as the end of a ngram or not). For each ngram segmentation with expected length I/2, the time complexity is $O((I/2)^3)$ using the efficient Hungarian algorithm. In this way, the total computational complexity of the original two conditions is $O(2^I I^3)$.

For computational tractability, we loosen the conditions by:
\begin{itemize}
    \item [1.] We consider all ngrams in the target sentence to avoid searching the ngram segmentation. In other words, each word is allowed to occur in multiple ngrams in one ordering $O$.
    \item [2.] We only consider ngrams with a fixed size $N$ (e.g., only bigrams), which enables us to cast this problem as Maximum Bipartite Matching and leverage the efficient Hungarian algorithm, as done in~\cite{Du:2021:ICML}.
\end{itemize}
By loosening the conditions, there are (I-N+1) ngrams of size $N$ in the sentence, and the computational complexity is $O(I^3)$.
Accordingly, the loss of the ordering $O^j$ is computed as:
\begin{eqnarray}
    P_G(O^{j}|X) &= \prod\limits_{y_{i:i+N-1} \in O^j} P_G(y_{i:i+N-1}|X) .
\end{eqnarray}

Figure~\ref{fig:ngram-oaxe} shows the calculation of bigram-\oaxe loss for the target sentence ``I ate pizza this afternoon''. We consider all bigrams in the sentence (see ``Bigram List''), and obtain the probability distribution of the considered bigrams. 
We construct the bipartite graph $G = (U, V, E)$ where the first part of vertices $U$ is the set of N-1 neighbouring positions (e.g., the first two positions``Pos:1,2''), and the second part of vertices $V$ is the list of N-1 target bigrams. Each edge in E is the prediction log probability for the bigram in the corresponding position. We can follow~\newcite{Du:2021:ICML} to leverage the efficient Hungarian algorithm~\cite{kuhn1955hungarian} for fast calculation of \noaxe (see the assigned probabilities for the consider bigrams).

\label{sec:app-pseudocode}
\begin{algorithm}[t]
   \caption{Bigram-\oaxe Loss}
   \label{alg:pseudocode}
    \begin{algorithmic}
   \STATE {\bfseries Input:} Ground truth $Y$, NAT output $\log P$
   \STATE $bs$, $len$ = $Y$.size()
   \STATE $Y$ = $Y$.repeat(1, $len$).view($bs$, $len$, $len$)
   \STATE $costM$ = -$\log P$.gather(index=$Y$,~dim=2)
   \STATE {\color{red}$costM$ =~$costM$[:,~:-1,~:-1] +$costM$[:,~1:,~1:]}
  \FOR{$i=0$ {\bfseries to} $bs$}
  \STATE $bestMatch$[i] = HungarianMatch($costM$[i])
  \ENDFOR
   \STATE {\bfseries Return:}$costM$.gather(index=$bestMatch$)
\end{algorithmic}
\end{algorithm}

\paragraph{Implementation}
Algorithm~\ref{alg:pseudocode} shows the pseudo-code of \noaxe with $N=2$. 
The implementation of \noaxe is almost the same with that of \textsc{OaXE}, except that we add one more line (in red color) for constructing the probability distribution of ngrams. We implement \noaxe on top of the source code of \textsc{OaXE}, and leverage the same recipes (i.e., loss truncation and XE pretrain) to effectively restrict the free-order nature of \textsc{OaXE}.

Since both \noaxe and \oaxe only modify the training of NAT models, their inference latency is the same with the CMLM baseline (e.g., 15.3x speed up over the AT model).
Concerning the training latency, \oaxe takes 36\% more training time over the CMLM baseline, and our \noaxe takes 40\% more training time, which is almost the same to \oaxe since we only add one more line of code.

\paragraph{Discussion}
Some researchers may doubt that the \noaxe loss is not an intuitively understandable ``global'' loss, since some words are counted multiple times.

We use the example in Figure~\ref{fig:ngram-oaxe} to dispel the doubt. Firstly, except for the first and last words (i.e., ``I'' and ``afternoon''), the \noaxe loss equally counts the other words twice, which would not introduce the count bias.

Secondly, we follow~\newcite{Du:2021:ICML} to start with an initialization pre-trained with the XE loss, which ensures that the NAT models can produce reliable token probabilities to compute ngram probabilities. We also use the {\bf loss truncation} technique~\cite{losstruncation} to drop invalid ngrams with low probabilities (e.g., ``pizza this'' | Pos:2,3) in the selected ordering $O^j$.

Thirdly, the overlapped ngrams can help to produce more fluent translations by modeling global context in a manner of ngram LM. For example, the high-probability overlapped token in position 4 ``ate'' (i.e., P(ate | Pos:4) = 0.4) will guide NAT models to assign high probabilities to the neighbouring ngrams (``I ate'' | Pos:3,4) and (``ate pizza'' | Pos:4,5), which form a consistent clause (``I ate pizza | Pos:3,4,5''). 
In contrast, \noaxe would not simultaneously assign high probabilities to the phrases (``this afternoon'' | Pos:1,2) and (``pizza this'' | Pos:2,3), since the two phrases require NAT models to assign high probabilities to two different words (i.e., ``afternoon'' and ``pizza'') in the overlapped position 2.

The $\mathcal{L}_\text{\noaxe}$ loss with $N=2$ in Figure~\ref{fig:ngram-oaxe} is calculated as:
\begin{eqnarray*}
&& \log P(``this\ afternoon" | Pos:1,2) + \\
&& \log P(``I\ ate" | Pos:3,4) + \\
&& \log P(``ate\ pizza" | Pos:4,5)
\end{eqnarray*}
where the low-probability bigram (``pizza this'' | Pos:2,3) is truncated. In this way, \noaxe carries out operation at the ngram granularity: \noaxe requires exact match of the word order within the ngram phrases, and allows reordering between phrases (e.g., ``I ate pizza | Pos:3,4,5'' and ``this afternoon | Pos:1,2").

\section{Experiment}

\subsection{Experimental Setup}
\label{app:setup}

\paragraph{Data}
We conducted experiments on major benchmarking datasets that are widely-used in previous NAT studies~\cite{flowseq,imputer}: WMT14 English$\Leftrightarrow$German (En$\Leftrightarrow$De, 4.5M sentence pairs) and WMT17 English$\Leftrightarrow$Chinese (En$\Leftrightarrow$Zh, 20.0M sentence pairs).
We preprocessed the datasets with a joint BPE~\cite{Sennrich:BPE} with 32K merge operations for the En$\Leftrightarrow$De and En$\Leftrightarrow$Zh datasets. 
For fair comparison with prior work, we reported the Sacre BLEU~\cite{post2018call}\footnote{SacreBLEU hash : BLEU+case.mixed+lang.en-zh + numrefs.1+smooth.exp+test.wmt17+tok.zh+version.1.4.2} 
on the En-Zh task, and the compound BLEU~\cite{papineni2002bleu} on the other tasks.

\paragraph{Knowledge Distillation} 
We closely followed previous works on NAT to apply sequence-level knowledge distillation~\cite{kim2016sequence} to reduce the modes of the training data.
Specifically, we obtained {\bf distilled data} by replacing the target side of the original training data (i.e., {\bf raw data}) with translation produced by an external AT teacher.
Consistent with previous works~\cite{maskp,axe,Du:2021:ICML}, we employed Transformer-\textsc{Big}~\cite{transformer} as the AT teacher for knowledge distillation.

\paragraph{NAT Models}
We validated our approach on the representative NAT model -- CMLM~\cite{maskp}, which uses the conditional mask LM~\cite{devlin2019bert} to generate the target sequence from the masked input.
The NAT model shares the same architecture as Transformer-\textsc{Base}~\cite{Wang2020RethinkingTV}: 6 layers for both the encoder and decoder, 8 attention heads, 512 model dimensions.
We chose the CMLM models with the vanilla XE loss~\cite{maskp} and the \oaxe loss~\cite{Du:2021:ICML}
as our two main baselines.
To keep consistent with main baselines, we set 5 as length candidates for all CMLM models during inference.

We generally followed the hyperparameters used in~\cite{maskp}.
We trained batches of approximately 128K tokens using Adam~\cite{kingma2015adam}. 
The learning rate warmed up to $5\times10^{-4}$ in the first 10K steps, and then decayed with the inverse square-root schedule.
We trained all models for 300k steps, measured the validation BLEU at the end of each epoch, and averaged the 5 best checkpoints.
We followed~\cite{nathint,em,imputer,Du:2021:ICML} to use de-duplication trick to remove repetitive tokens in the generated output.

\iffalse
\subsection{Experimental Setup}
\label{sec:mt-setup}

\paragraph{Data}
We conducted experiments on several benchmarks with different data scales: WMT16 English$\Leftrightarrow$Romanian (En$\Leftrightarrow$Ro, 0.6M sentence pairs), WMT14 English$\Leftrightarrow$German (En$\Leftrightarrow$De, 4.5M), WMT17 English$\Leftrightarrow$Chinese (En$\Leftrightarrow$Zh, 20.0M), and WMT20 En$\Leftrightarrow$De dataset (45.1M).
We learned a BPE model~\cite{Sennrich:BPE} with 32K merge operations for the En$\Leftrightarrow$De and En$\Leftrightarrow$Zh datasets, and 40K merge operations for the the En$\Leftrightarrow$Ro dataset.
For fair comparison with prior work, we reported the Sacre BLEU~\cite{post2018call}
on the En-Zh task, and BLEU score~\cite{papineni2002bleu} on the other tasks.

\paragraph{NAT Models}
We validated our approach on the representative NAT model -- CMLM~\cite{maskp}, and reproduced the reported results using the standard hyper-parameters. 
We employed Transformer-\textsc{Big}~\cite{transformer} for distilling the En$\Leftrightarrow$De and En$\Leftrightarrow$Zh datasets, and Transformer-\textsc{Base} for distilling the En$\Leftrightarrow$Ro dataset.
We chose the CMLM models with the vanilla XE loss~\cite{maskp} and the \oaxe loss~\cite{Du:2021:ICML} as our two main baselines.
We followed~\cite{nathint,em,imputer,Du:2021:ICML} to use de-duplication trick to remove repetitive tokens in the generated output.

Please refer to Appendix~\ref{app:setup} for more details.
\fi

\begin{figure}[t]
    \centering 
    \subfloat[Raw Data: En-De]{
    \includegraphics[height=0.3\textwidth]{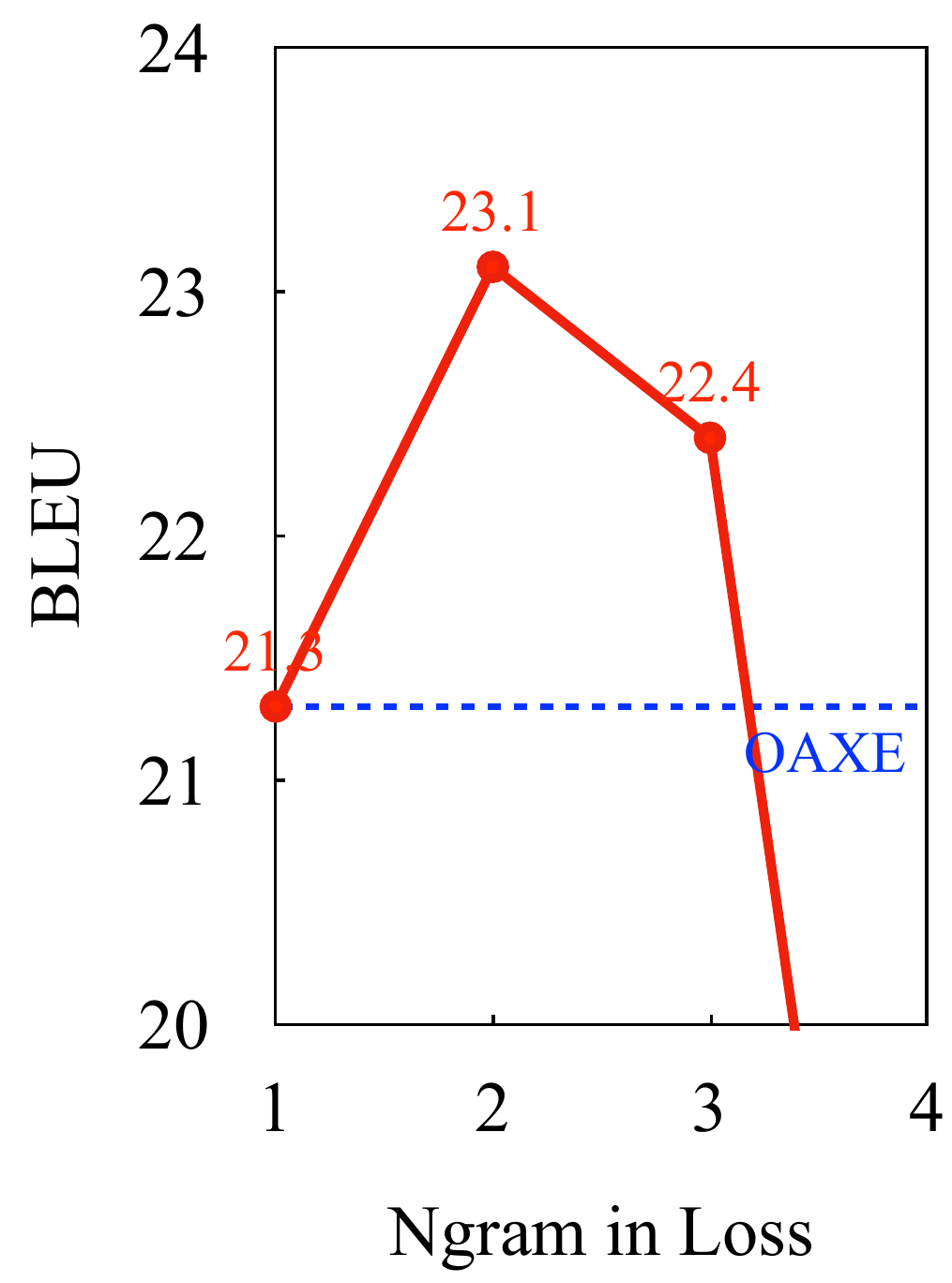}}
    \hfill
    \subfloat[Raw Data: De-En]{
    \includegraphics[height=0.3\textwidth]{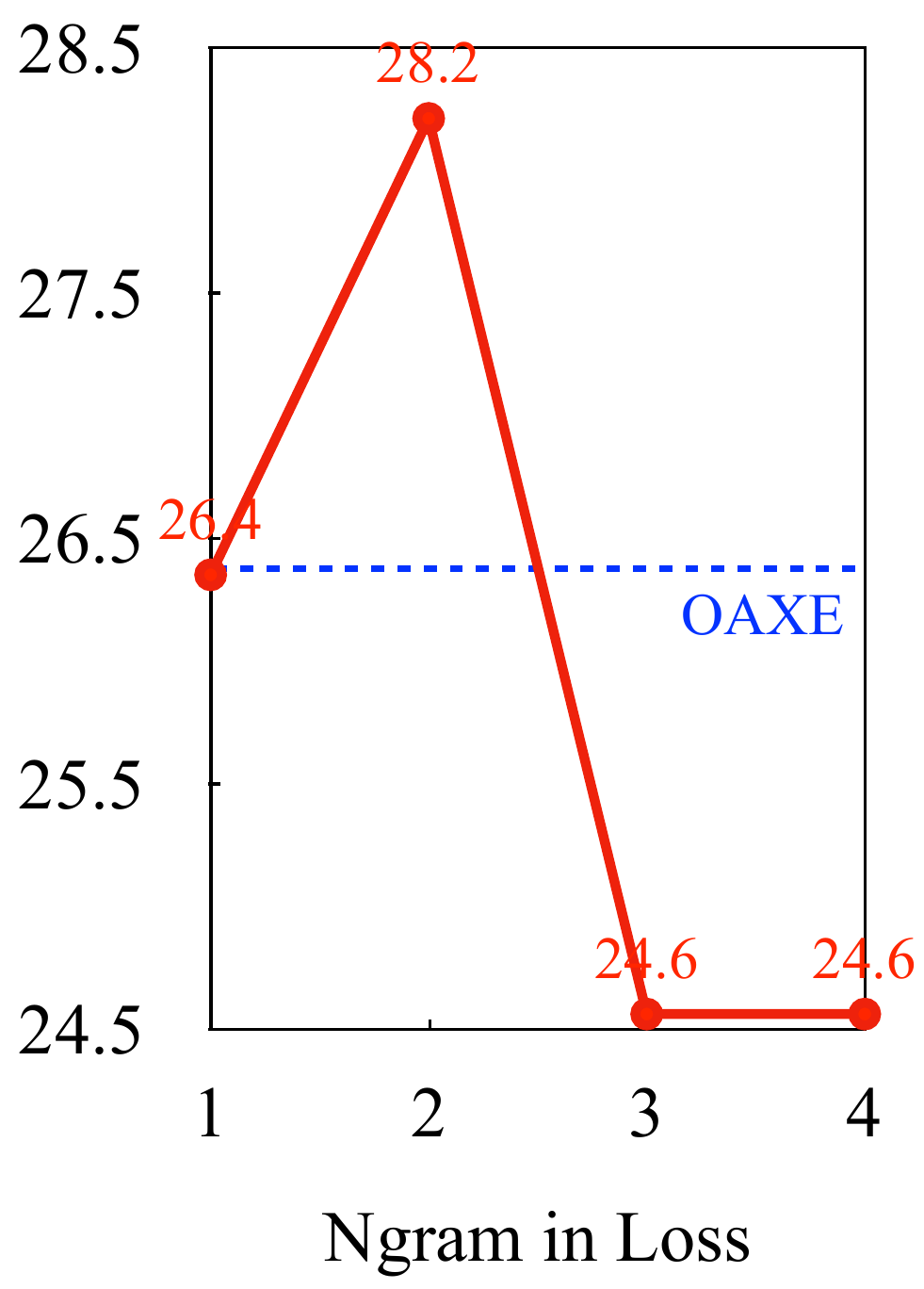}}
    \subfloat[Distilled Data: En-De]{
    \includegraphics[height=0.3\textwidth]{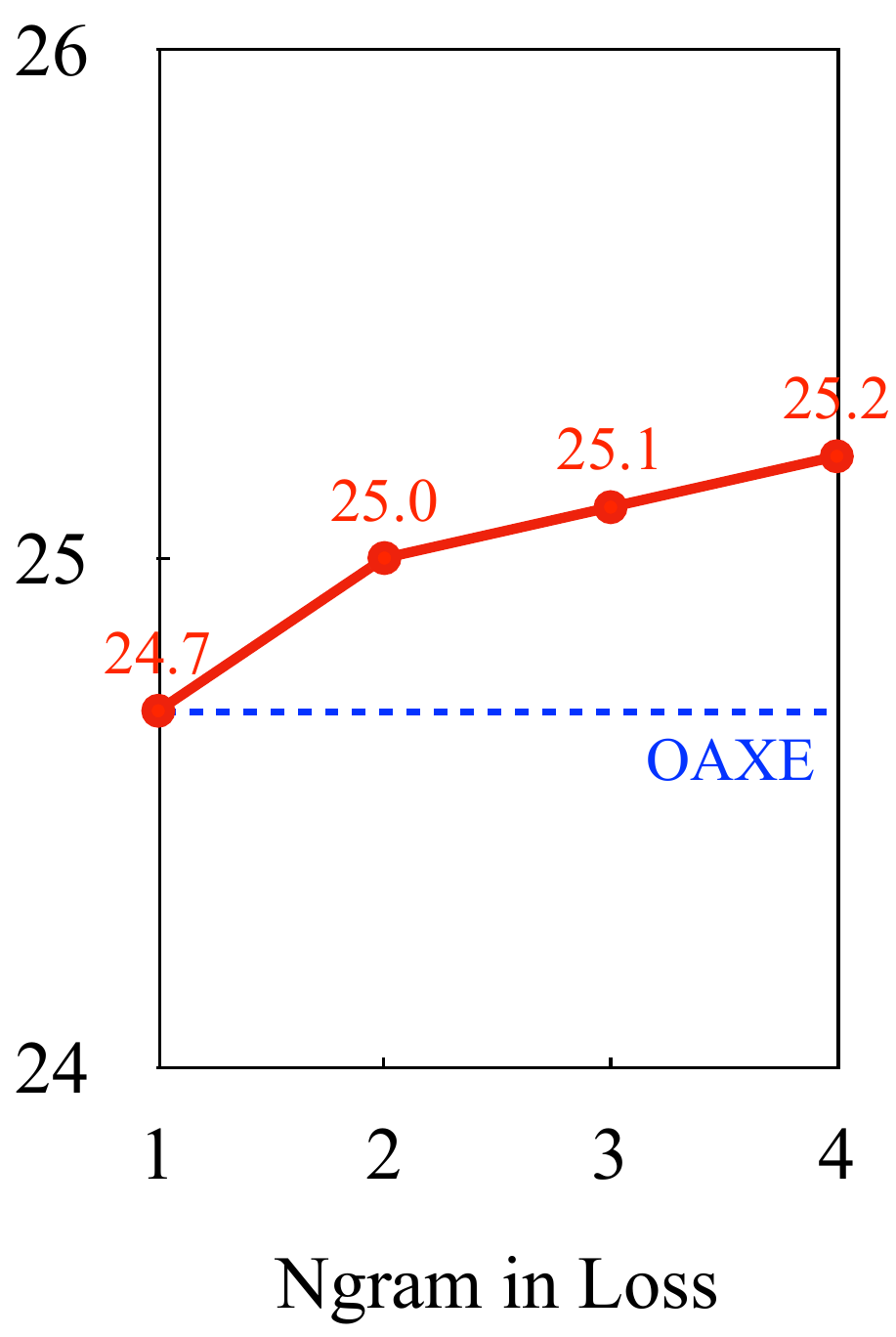}}
    \hfill
    \subfloat[Distilled Data: De-En]{
    \includegraphics[height=0.3\textwidth]{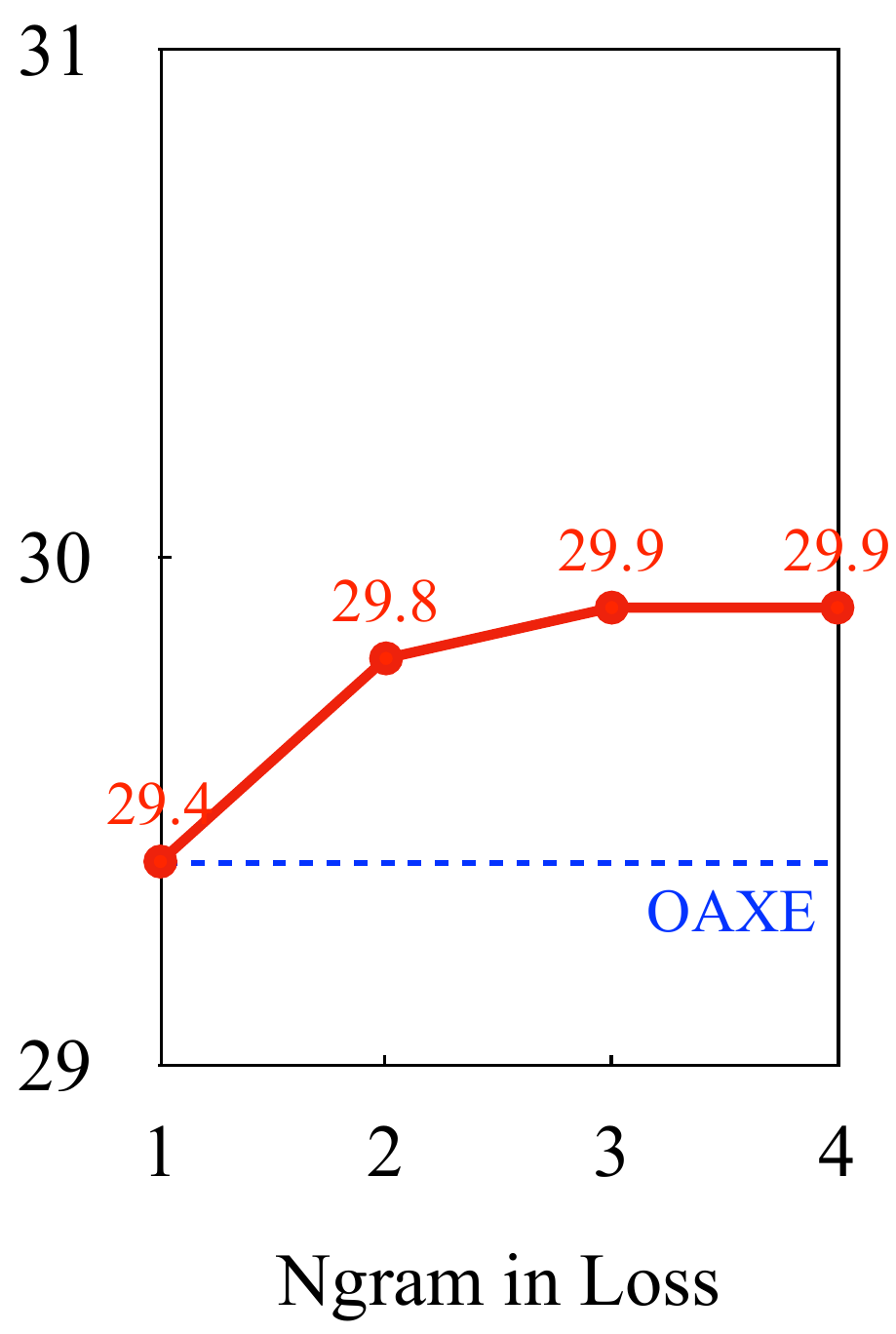}}
    \caption{Impact of N-gram choice in \noaxe loss (i.e., $N$ in Equation~\ref{eqn:ngram-distributions}). % on the WMT14 En$\Leftrightarrow$De validation sets. 
    \oaxe can be viewed as a special case of \noaxe with $N=1$.}

    \label{fig:ablation-choice-n}
\end{figure}

\subsection{Ablation Study}

In this section, we investigated the impact of different components for \noaxe
on the WMT14 En$\Leftrightarrow$De validation sets.

\paragraph{Impact of Ngram Size}
We first investigated the impact of different $N$ in the \noaxe loss on the translation performance.
Figure~\ref{fig:ablation-choice-n} shows the results for both raw data and distilled data.
As seen, the bigram-\textsc{OaXE} achieves the best performance on raw data, while 4gram-\textsc{OaXE} performs best on the distilled data. We attribute the different behaviors to the difficulty of the dataset: raw data contains more modes than distilled data~\cite{NAT}, thus it is more difficult to learn larger ngrams from the complicated raw data. In the following experiments, we set $N=2$ for raw data, and $N=4$ for distilled data.

\begin{figure}[t]
    \centering
    \centering 
    \subfloat[Distilled Data: En-De]{
    \includegraphics[height=0.33\textwidth]{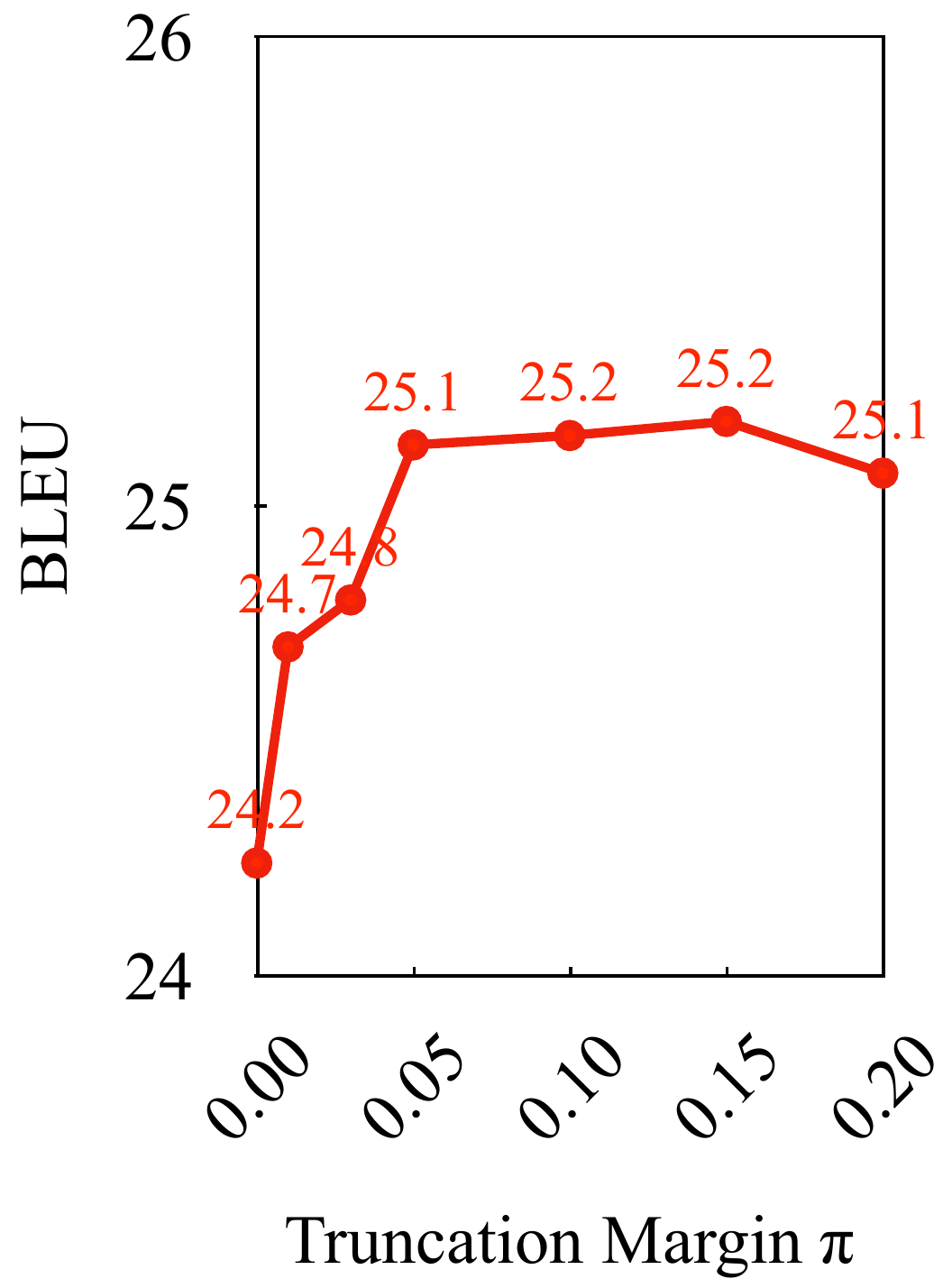}}
    \hfill
    \subfloat[Distilled Data: De-En]{
    \includegraphics[height=0.33\textwidth]{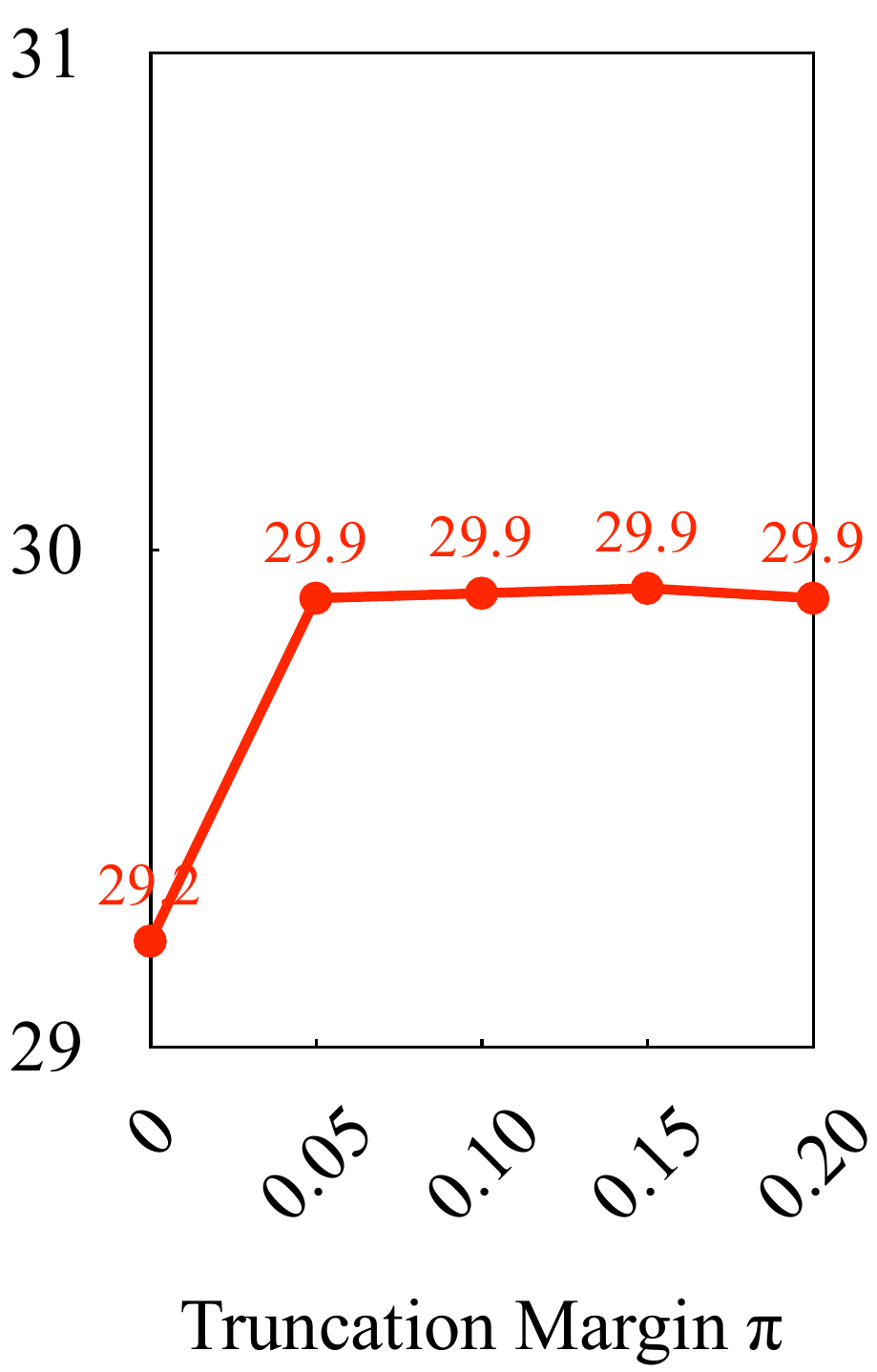}}
    \caption{Impact of the truncation margin $\pi$ for NAT models trained on the distilled data.}
    \label{fig:impact-of-truncation}
\end{figure}

\paragraph{Impact of Truncation Margin}
\label{sec: pretrain}
Figure~\ref{fig:impact-of-truncation} shows the impact of truncation margin $\pi$, which is searched from \{$0, 0.05, 0.10, 0.15, 0.20$\}.
Intuitively, higher $\pi$ drops more ngrams. When $\pi$ increases from 0 to 0.05, we achieved 0.7$\sim$0.9 BLEU improvement by dropping likely invalid ngrams.
\noaxe is robust to the truncation margin $\pi$: when $\pi$ further increases, the performance does not vary too much. We follow~\newcite{Du:2021:ICML} to use $\pi=0.15$ for all language pairs and datasets in the following experiments.

\subsection{Translation Performance}

In this section, we conduct comprehensive experiments to validate the effectiveness of %{\em ngram}-\textsc{OaXE}.
the proposed \noaxe model. 
First, we use multiple semantically equivalent references to better evaluate the multimodality nature of generated translation, which serves as the main results for analyses in the following sections.
Then we compare our approach with previous work on the benchmarking testsets with single reference.

\paragraph{Multiple References}

We follow~\newcite{Du:2021:ICML} to use two test sets with multiple references: 1) the dataset released by~\newcite{ott2018analyzing} that consists of ten human translations for 500 sentences taken from the WMT14 En-De test set; and 2) the combination of NIST02-08 Zh-En test sets that consists of 7497 sentences with four references. The translation models are trained on the WMT14 En-De and WMT17 Zh-En training data, respectively.

\begin{table}[!t]
    \centering
    \setlength{\tabcolsep}{3pt}
    \begin{tabular}{lllll}
    \toprule
    \multirow{2}{*}{\bf Model}   &  \multicolumn{2}{c}{\bf W14 En-De}   &  \multicolumn{2}{c}{\bf NIST Zh-En}\\
    \cmidrule(lr){2-3} \cmidrule(lr){4-5}
        &  \bf BLEU & \bf $\Delta$    &  \bf BLEU & \bf $\Delta$\\
    \midrule
    \bf Raw Data  \\
    Transformer & 71.4  & - & 41.7 & -\\
    \hdashline
    CMLM        &  28.1 & - & 12.1 & -\\
    ~~+\oaxe    &  57.5 & +29.4 & 36.5 & +24.4\\
    ~~+\noaxe   &  61.3$^{\uparrow\Uparrow}$ & +33.2 & 38.6$^{\uparrow\Uparrow}$ & +26.5\\
    \midrule     
    \bf Distilled Data  \\
    Transformer & 72.7 & - & 42.0 & -\\
    \hdashline
    CMLM        &  50.7 & - & 23.7 & - \\
    ~~+\oaxe    &  68.0 & +17.3 & 40.4 & +16.7\\
    ~~+\noaxe   &  68.9 $^{\uparrow\Uparrow}$ &+18.2 & 41.2 $^{\uparrow\Uparrow}$ &+17.5 \\
    \bottomrule
    \end{tabular}
    \caption{BLEU scores on test sets with multiple references. ``$\Delta$'' denotes the improvement over CMLM. ``$\uparrow$'' and ``$\Uparrow$'' denotes significantly better than CMLM and \oaxe with $p < 0.05$, respectively. The Zh-En NMT model is trained on the WMT17 Zh-En data.}
    \label{tab:reference}
\end{table}

\begin{table*}[t]
    \centering
    \begin{tabular}{l ll}
    \toprule
    \multirow{2}{*}{\bf Model} & \multicolumn{2}{c}{\textbf{WMT14}}\\
    \cmidrule(lr){2-3}
     & \textbf{En-De} &\textbf{De-En}\\
    \midrule
    {\bf Autoregressive} Transformer &   27.6 & 31.4\\
    \midrule
    \bf Non-Autoregressive \\
    CTC Loss~\cite{libovicky-helcl-2018-end}   & 17.7 & 19.8\\
    Flowseq~\cite{flowseq}                     & 18.6 & 23.4\\
    Imputer~\cite{imputer}                     & 15.6 & -\\
    \hdashline
    CMLM~\cite{maskp}                          & 10.6 & 15.1\\
    ~~+AXE~\cite{axe}                          & 20.4 & 24.9\\
    ~~+Correction~\cite{huang2022improving}    & 20.6 & 25.4\\
    ~~+\oaxe~\cite{Du:2021:ICML}               & 22.4 & 26.8\\
    ~~+\noaxe({\em Ours})  & \bf 23.6$^{\uparrow\Uparrow}$ & \bf 27.9$^{\uparrow\Uparrow}$ \\
    \bottomrule
    \end{tabular}
\caption{BLEU scores on testsets with single reference for NAT models trained on the {\bf raw data}.} 
\label{tab:en-de-raw}
\end{table*}

\begin{table*}[t]
\centering
\begin{tabular}{l ll ll}
\toprule
\multirow{2}{*}{\bf Model} & \multicolumn{2}{c}{\textbf{WMT14}} & \multicolumn{2}{c}{\textbf{WMT17}} \\
 \cmidrule(lr){2-3}\cmidrule(lr){4-5}
 & \textbf{En-De} &\textbf{De-En} &\textbf{En-Zh}  &  \textbf{Zh-En}\\
\midrule
% \bf {Autoregressive} \\
{\bf Autoregressive} Transformer   &  27.8 & 31.3 &  34.4 & 24.0\\
\midrule
\bf Non-Autoregressive \\
Bag-of-ngrams~\cite{natbow}  & 20.9 & 24.6 & - & -\\
Flowseq~\cite{flowseq}            & 21.5 & 26.2 & - & -\\
Bigram CRF~\cite{natcrf}     & 23.4 & 27.2 & - & -\\
Imputer~\cite{imputer}            & 25.8 & 28.4 & - & -\\
\hdashline
CMLM~\cite{maskp}                 & 18.1 & 21.8 & 24.2 & 13.6\\
~~+AXE~\cite{axe}               & 23.5 & 27.9 & 30.9 & 19.8\\
~~+GLAT~\cite{glat}     & 25.2 & 29.8 & - & -\\
~~+CTC+VAE~\cite{gu-kong-2021-fully} & 27.5 & 31.1 & - & -\\
~~+\oaxe~\cite{Du:2021:ICML}  & 26.1 & 30.2 & 32.9 & 22.1\\
~~+\noaxe ({\em Ours})        & \bf 26.5$^{\uparrow\Uparrow}$ & \bf 30.5$^{\uparrow\Uparrow}$ & \bf 33.2$^{\uparrow\Uparrow}$ & \bf 22.8$^{\uparrow\Uparrow}$\\
\bottomrule
\end{tabular}
\caption{BLEU scores on testsets with single reference for NAT models trained on the {\bf distilled data}.
}
\label{tab:main}
\end{table*}

Table~\ref{tab:reference} lists the translation performance.
Encouragingly, \noaxe narrows the performance gaps between:
\begin{itemize}[leftmargin=10pt]
    \item {\em NAT models trained on raw data and on distilled data}: Take W14 En-De as an example, knowledge distillation brings an improvement of 22.6 BLEU points over raw data for XE (i.e., from 28.1 to 50.7). \oaxe narrows the gap to 10.5 BLEU points (i.e., 57.5 vs. 68.0), and our \noaxe further narrows the gap to 7.6 BLEU points (i.e., 61.3 vs. 68.9), moving toward training NAT models without distillation.
    \item {\em NAT and AT models trained on raw data}: For independent NAT models without distillation, \oaxe reduces the performance gap from 43.3 to 13.9 for En-De, and from 29.6 BLEU to 5.2 BLEU for Zh-En. {\em Ngram}-\oaxe further reduces the gaps to 10.1 and 3.1 BLEU points, indicating the potential of NAT to become a practical system without relying on external resources.
\end{itemize}

\paragraph{Benchmarks with Single Reference}
We also evaluated the performance of fully NAT models on benchmarks with single reference.
In addition to the closely related XE variants (e.g., AXE and \textsc{OaXE}), %~\cite{axe,Du:2021:ICML}, 
we also compare against several strong baseline models: 
1) CTC Loss -- a NAT model with latent alignments~\cite{libovicky-helcl-2018-end};
2) Flowseq -- a latent variable model based on generative flow~\cite{flowseq}; 
3) Imputer -- an extension of CTC with the use of distillation during training~\cite{imputer};
4) Corretion -- a NAT model with error correction mechanism~\cite{huang2022improving};
5) Bigram CRF -- the CRF-based semi-autoregressive model~\cite{natcrf}; 
6) GLAT -- Glancing-based training~\cite{glat}; 
7) CTC+VAE -- combining the CTC loss and latent variables (VAE)~\cite{gu-kong-2021-fully}.

Table~\ref{tab:en-de-raw} lists the results on the {\bf raw data} % without distillation.
that do not rely on any external resources (e.g., AT models for KD).
CMLMs trained by \noaxe improves over the XE-trained baseline by 12.8 BLEU points on average, and outperforms the strong \oaxe by +1.0 BLEU points. These results indicate that the ngram supervision helps to better capture the complicated patterns from the raw data.

Table~\ref{tab:main} lists the BLEU scores on the {\bf distilled data}. Our approach consistently improves over the strong \textsc{OaXE} loss in all cases, demonstrating the effectiveness and universality of the proposed \noaxe loss.
Our approach also outperforms all existing NAT using a single technique (exclude~\newcite{gu-kong-2021-fully} with two techniques).

\section{Analysis}

In this section, we provide some insights where \noaxe improves over CMLM (i.e., XE) and \oaxe from different perspectives. Otherwise stated, we report results on the WMT14 En-De test set with multiple references (i.e., the column ``W14 En-De'' in Table~\ref{tab:reference}).

\subsection{Analysis of Ngram Translation}

\begin{table}[t]
    \centering
    \begin{tabular}{lrrrr}
    \toprule
    \multirow{2}{*}{\bf Model}   &   \multicolumn{4}{c}{\bf Ngram Size}\\
    \cmidrule(lr){2-5}
        &   1   &   2   &   3 & 4\\
    \midrule
    \multicolumn{5}{l}{\bf Raw Data}\\
    CMLM    & 81.8 &48.8 &30.1 &20.4\\
    ~~~+\oaxe   & 86.0 &64.3 &48.0 &35.6\\
    ~~~+\noaxe  & 88.2 &69.8 &54.7 &42.3\\
    \midrule
    \multicolumn{5}{l}{\bf Distilled Data}\\
    CMLM    &86.0 &63.3 &47.1 &36.0 \\
    ~~~+\oaxe   &90.6 &75.4 &62.0 &51.1 \\
    ~~~+\noaxe  &90.5 &76.1 &62.9 &52.0 \\
    \bottomrule
    \end{tabular}
    \caption{Accuracy (\%) of the generated ngram phrases in the model outputs.}
    \label{tab:ngram-bleu}
\end{table}

\begin{table}[t]
\small
    \centering
    \begin{tabular}{c m{5.8cm}}
    \toprule
    \bf Refer. & The Vollmaringen Male Voice Choir got things running with atmospheric {\color{blue} songs such as} ``{\color{blue} Im Weinparadies}'' and ``{\color{blue} Lustig, ihr Brüder}''.\\
    \midrule
    \bf CMLM &The MGV Vollmaringen opened with atmospheric {\color{red} songs songs as} ``{\color{red} Im Weinparapara}'' and ``{\color{red} Lustig, her brothers}''.\\
    \hdashline
    \bf \oaxe &The MGV Vollmaringen opened with atmospheric {\color{red} songs such ``''} ``{\color{red} Im Weinpara}'' and ``{\color{red} Lustig, her brothers}''.\\
    \hdashline
    \bf Ours & The MGV Vollmaringen opened with atmospheric {\color{blue} songs such as} ``{\color{blue} Im Weinparadies}'' and ``{\color{blue} Lustig, ihr Brüder}''\\
    \bottomrule
    \end{tabular}
    \caption{Examples of De-En translation for NAT models trained on distilled data. \oaxe model often mistakenly translates some ngram phrases (in {\color{red}{red}} color), 
    and \noaxe can correctly translate them (in {\color{blue}{blue}} color).}
    \label{tab:case-ngram}
\end{table}

We first investigate whether the proposed \noaxe improves the generation of phrases in the output. To this end, we use the individual ngram scores as the accuracy of generating ngrams of the corresponding size. An individual ngram score is the evaluation of just matching ngrams of a specific size, such as unigram and bigram.\footnote{ The individual ngram BLEUs are generally produced by the BLEU script for the same model outputs, and do not refer to the model training. For example, given the output of \noaxe trained on the WMT14 En-De raw data, the script outputs ``$BLEU = 61.3\ 88.2/69.8/54.7/42.3$''. The final BLEU score is 61.3, and the individual 1gram, 2gram, 3gram, and 4gram BLEU scores are 88.2, 69.8, 54.7, and 42.3, respectively.}
As shown in Table~\ref{tab:ngram-bleu}, our \noaxe consistently outperforms the \oaxe counterparts in all ngram levels and the improvement goes up with the increase of ngram, demonstrating that \noaxe indeed raises the ability of NAT model on capturing the patterns of ngram phrases.

\paragraph{Case Study}
Table~\ref{tab:case-ngram} shows an translation example on the WMT14 De-En testset.
The vanilla CMLM model mistakenly generates the ngram phrase ``{\em songs songs as}'' with repeated words ``{\em songs}". Although the \oaxe model remedies the repetition problem, it fails to generate the phrase ``{\em such as}''. In addition, both the CMLM and \oaxe models fail to generate the names of the two songs ``{\em Im Weinparadies}'' and ``{\em Lustig, ihr Brüder}''. Our \noaxe successfully generate all the three ngram phrases.

\subsection{Analysis of Structure Modeling}
\label{sec:structure-order}

\begin{figure}[t]
  \centering
  \begin{subfigure}[b]{0.4\textwidth}
  \centering
  \includegraphics[width=\textwidth]{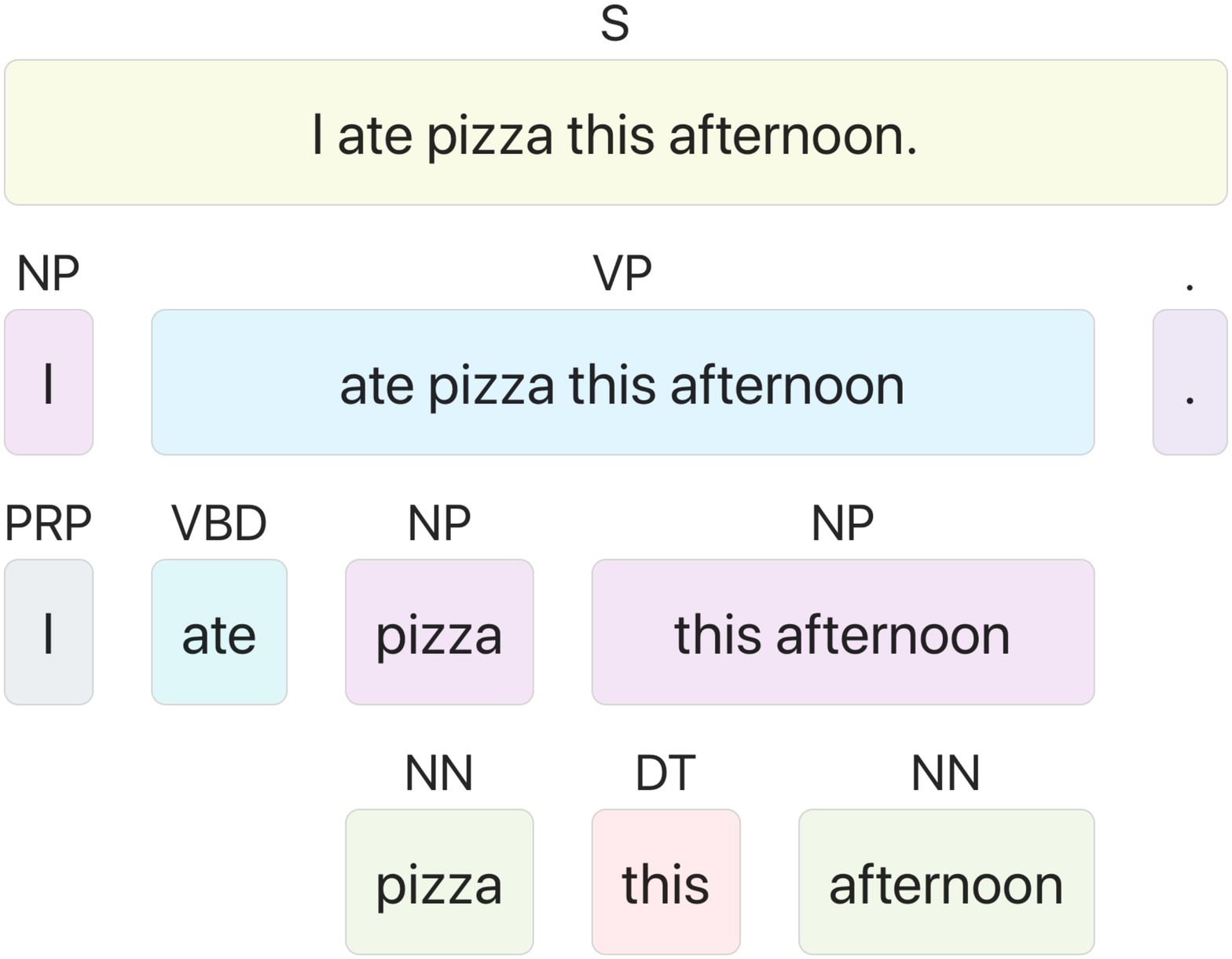}
  \caption{An example constituent tree.}
  \label{fig:constituent-tree}
  \end{subfigure}
  \\
   \begin{subtable}[b]{0.45\textwidth}
   \centering
   \begin{tabular}{c l}
    \toprule
    \bf Level   &   \bf Syntactic Sequence\\
    \midrule
    3   &   NP  VP  .\\
    2   &   PRP VBD NP NP .\\
    1   &   PRP VBD NN DT NN .\\
    \bottomrule
    \end{tabular}
    \caption{Syntactic sequences at different levels.} 
    \label{fig:syntactic-sequence}
   \end{subtable}
%   }
  \caption{Constituent tree of the sentence ``I ate pizza this afternoon.'' (a), and the corresponding syntactic sequences at different levels (b, from bottom to up). ``2-level'' denotes the syntactic sequence at the last but one level of the constituent tree.}
\end{figure}

\iffalse
\begin{figure*}[h]
  \centering
  \begin{subfigure}[b]{0.4\textwidth}
  \centering
  \includegraphics[width=\textwidth]{figures/constituent_tree.pdf}
  \caption{An example constituent tree.}
  \label{fig:constituent-tree}
  \end{subfigure}
  \hspace{0.05\textwidth}
   \begin{subtable}[b]{0.45\textwidth}
   \centering
   \begin{tabular}{c l}
    \toprule
    \bf Level   &   \bf Syntactic Sequence\\
    \midrule
    3   &   NP  VP  .\\
    2   &   PRP VBD NP NP .\\
    1   &   PRP VBD NN DT NN .\\
    \bottomrule
    \end{tabular}
    \caption{Syntactic sequences at different levels.} 
    \label{fig:syntactic-sequence}
   \end{subtable}
  \caption{Constituent tree of the sentence ``I ate pizza this afternoon.'' (a), and the corresponding syntactic sequences at different levels (b, from bottom to up). ``2-level'' denotes the syntactic sequence at the last but one level of the constituent tree.}
\end{figure*}
\fi

\paragraph{Structure Ordering}
To assess the models' abilities of modeling reordering between ngram phrases, we follow~\newcite{Wang:2021:ACL} to measure the precision of outputs at the syntactic level, which can reflect the structure ordering at phrase level.
Specifically, we use the syntactic sequence at a certain layer  (Figure~\ref{fig:syntactic-sequence}) of the constituent tree of the generated outputs (Figure~\ref{fig:constituent-tree}). Generally, each tag at a higher syntactic level covers more words (e.g., ``VP (ate pizza this afternoon)'' at 3-level) and corresponds well to a ngram phrase. Accordingly, higher-level syntactic sequences denotes structure ordering at a larger granularity.
We calculate the BLEU score for the syntactic sequences of models' outputs to measure the precision of the structure ordering at different granularities.

\begin{table}[t]
    \centering
    \setlength{\tabcolsep}{4pt}
    \begin{tabular}{lllll}
    \toprule
    \multirow{2}{*}{\bf Model}   &   \multicolumn{4}{c}{\bf Syntactic Level}\\
    \cmidrule(lr){2-5}
        &  \bf 1   &  \bf 2   & \bf 3 &  \bf 4\\
    \midrule
    \multicolumn{5}{l}{\bf Raw Data}\\
    CMLM    & 18.3 & 11.5 &13.4  & 17.4 \\
    \oaxe   & 37.3  &28.5 &26.6 & 28.5 \\
    \noaxe  & 39.9$^{\uparrow\Uparrow}$ &30.9$^{\uparrow\Uparrow}$  &29.4$^{\uparrow\Uparrow}$ & 30.0$^{\uparrow\Uparrow}$\\
    \midrule
    \multicolumn{5}{l}{\bf Distilled Data}\\
    CMLM    & 31.5   &14.7 & 23.2 & 25.8 \\
    \oaxe   & 41.9  &33.0  & 30.4 & 31.2 \\
    \noaxe  & 42.6$^{\uparrow\Uparrow}$  &33.8$^{\uparrow\Uparrow}$  &31.4$^{\uparrow\Uparrow}$ & 32.3$^{\uparrow\Uparrow}$ \\
    \bottomrule
    \end{tabular}
    \caption{BLEU scores of the syntactic sequence at different levels (from bottom to up).} 
    \label{tab:syntactic-bleu}
\end{table}

Table~\ref{tab:syntactic-bleu} shows the results for syntactic sequences at different levels.
\oaxe significantly improves the precision of structure order over the CMLM baseline by a large margin, which is consistent with the claim of~\newcite{Du:2021:ICML} that \oaxe is better at modeling word order. Our \noaxe can further improve the precision of structure order, which we attribute to that \noaxe models ordering at a larger granularity (i.e., ngrams).

\begin{figure}[t]
    \centering 
    \subfloat[Raw Data]{
    \includegraphics[height=0.35\textwidth]{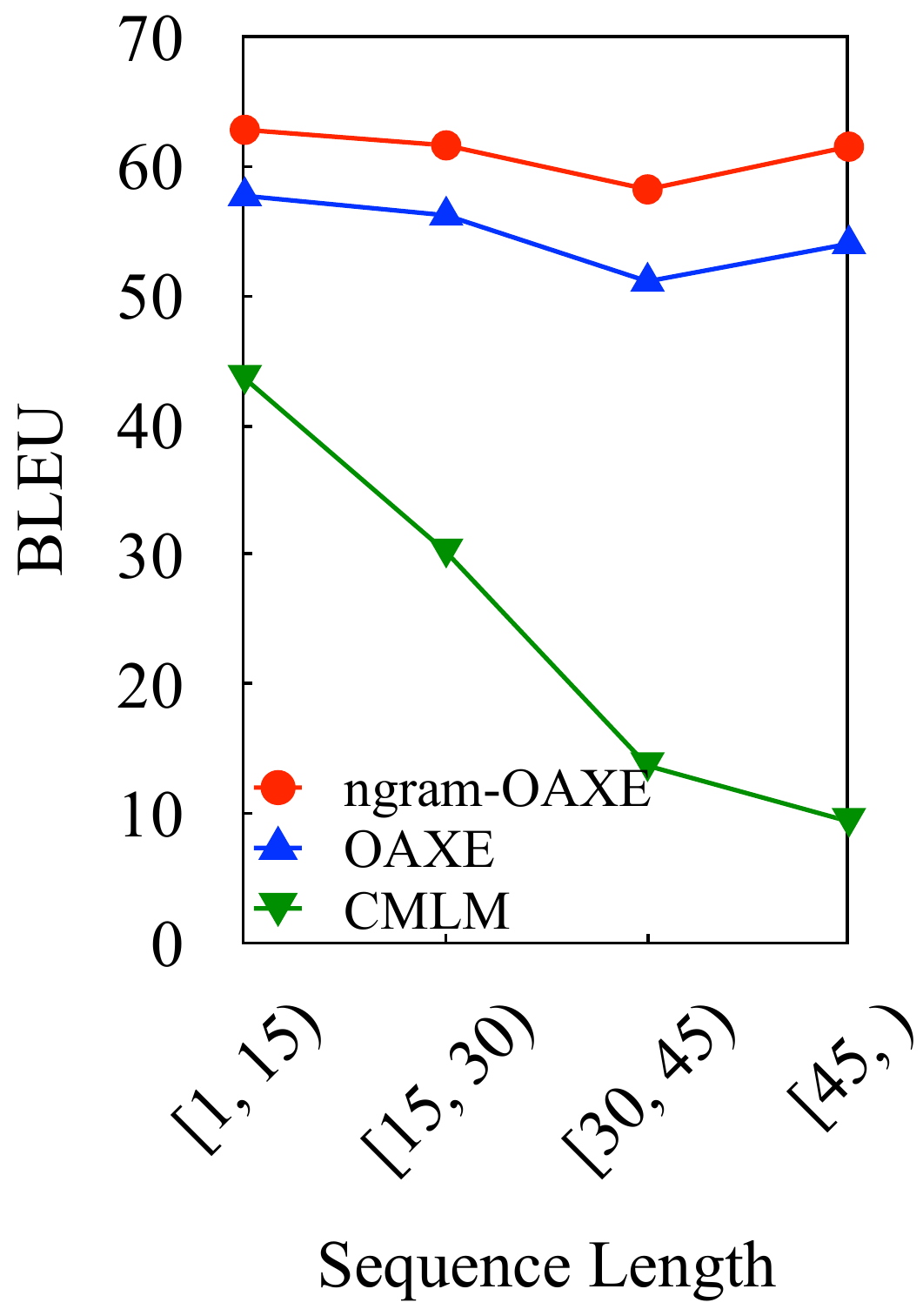}}
    \hfill
    \subfloat[Distilled Data]{
    \includegraphics[height=0.35\textwidth]{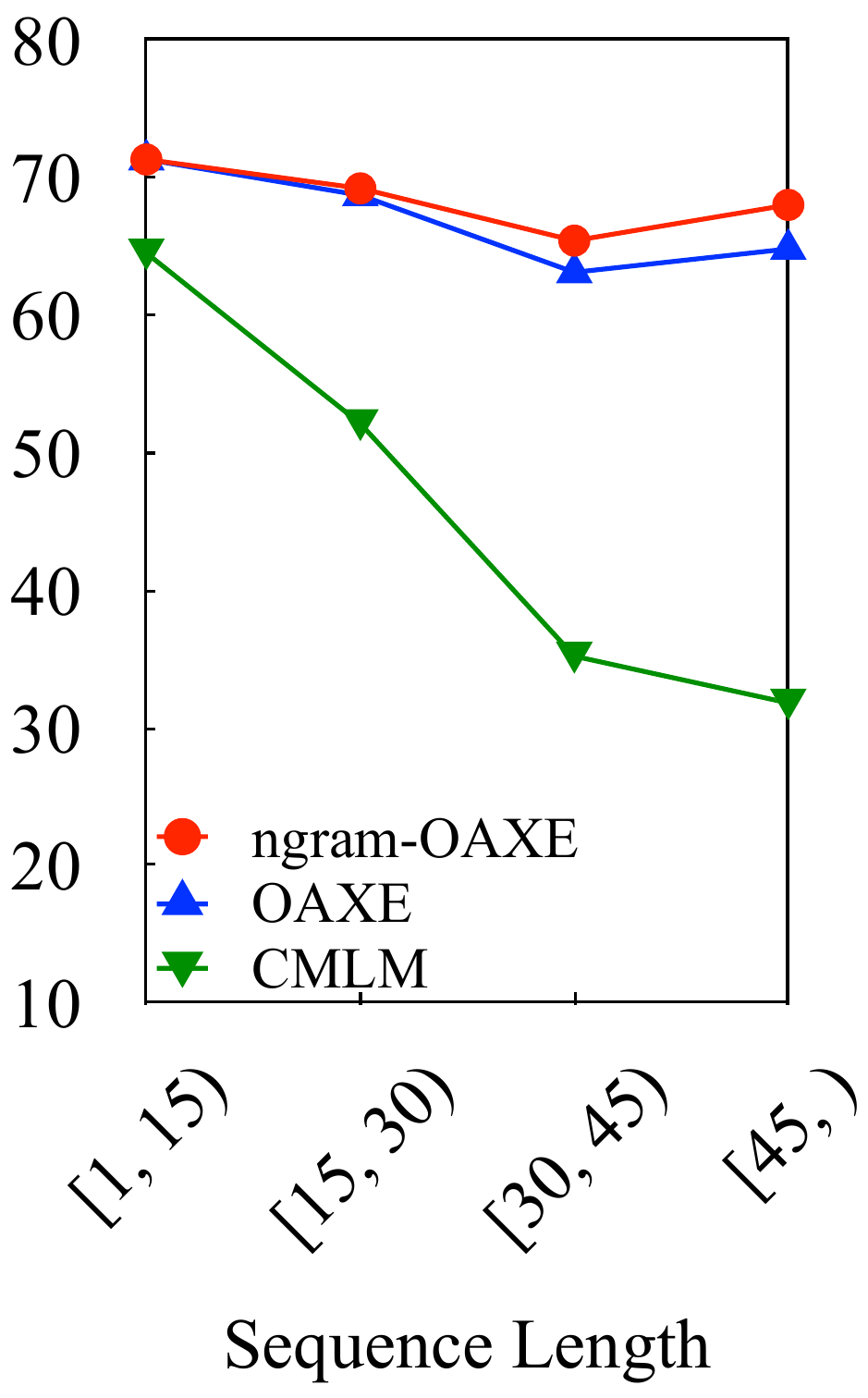}}
    \caption{Translation performance with respect to the length of the target sentence.}
    \label{fig:length}
\end{figure}

\paragraph{Sequence Length}
We also investigate the model performance for different sequence lengths. We split the test sets into different buckets based on the reference sentence length, indicating whether a system does better or worse at shorter or longer sentences. Generally, longer sentences are more complex in linguistic structure.
Figure~\ref{fig:length} shows that results on the sampled WMT14 En-De test set with multiple references. As seen, the performance of XE drops rapidly when the sequence length increases, and \oaxe can significantly improves performance on longer sentences with a better modeling of word order. Our \noaxe can handle long sequences even better, which we attribute to the strength of \noaxe on both translating longer ngrams and modeling structure ordering between ngram phrases.

\subsection{Analysis of Generated Output}
\label{sec:analysis-output}

\begin{table}[t]
    \centering
    \begin{tabular}{l rr}
    \toprule
    {\bf Model}  &  {\bf Repetition}  &  {\bf PPLs}\\
    \midrule
    \bf Gold Test Set   &   0.04\% &  90.9\\
    \midrule
    \bf Raw Data  \\
    CMLM          &   31.11\%   &   1820.2\\
    ~~+\oaxe      &    3.14\%   &   237.6\\
    ~~+\noaxe     &    2.99\%   &   199.9\\
    \midrule
    \bf Distilled Data\\
    CMLM          &12.10\% &1435.5\\
    ~~+\oaxe      &1.56\% &240.8\\
    ~~+\noaxe     &0.98\% &125.9\\
    \bottomrule
    \end{tabular}
    \caption{Analyses of the generated outputs.
    Lower repeated token percentage (``Repetition") denotes lower multimodality in a model. Lower perplexities (``PPLs") denote better fluency.
    }
    \label{tab:analysis-outputs}
\end{table}

\paragraph{Token Repetition}

One widely-cited weakness of existing NAT models is the multimodality problem, in which a model may consider many possible translations at the same time due to the independent predictions of target words~\cite{NAT}.
Accordingly, the NAT output typically contains many repetitive tokens (e.g., ``songs songs'' in Table~\ref{tab:case-ngram}).
We followed the common practices to use repeated token percentage for measuring multimodality in a NAT model, as listed in Table~\ref{tab:analysis-outputs}.
While \oaxe can mostly alleviate the repetition problem, the proposed \noaxe can further reduce the repeated percentage over the very strong baseline (e.g., 0.98\% vs. 1.56\% on distilled data).

\paragraph{Generation Fluency}
We followed~\newcite{Du:2021:ICML} to measure the generation fluency with language models released by Fairseq,\footnote{\url{https://github.com/pytorch/fairseq/blob/master/examples/language\_model/}}
which are trained on the News Crawl corpus for the target language. To better evaluate the fluency of the generated output, we use a practical trick {\em de-duplication}~\cite{imputer} to remove the repetitive tokens.
Clearly, \noaxe consistently improves fluency in all settings compared with \textsc{OaXE}. We attribute the fluency improvement to the strength of \noaxe on both translating longer ngrams\footnote{The longer ngrams generally account for the fluency of the translation.} and modeling sentence structures.

\section{Related Work}

\paragraph{Alleviating Multimodality Problem for NAT}

A number of recent efforts have explored ways to improve the NAT models' ability to handle multimodality. One thread of work iteratively refines the generated outputs with K decoding passes~\cite{iterativerefine,gu2019levenshtein}, which sacrifices the primary benefit of NAT models -- fast inference~\cite{kasai2021deep}.
To maintain the advantage of decoding efficiency, another thread of research aims to improve fully NAT models by building dependencies between target tokens~\cite{flowseq,Shu2020LaNMT}, or improving the training loss to ameliorate the effect of multimodality~\cite{axe,imputer,Du:2021:ICML}.

Knowledge distillation~\cite{kim2016sequence} is the preliminary step for the majority of NAT systems, which can effectively alleviate the multimodality problem by simplifying the training data~\cite{zhou2019understanding} and reducing the token dependency in target sequence~\cite{ren2020astudy}.
However, knowledge distillation relies on an external AT teacher, which prevents NAT models from self-completion. The ultimate goal is to train NAT models from scratch~\cite{huang2022improving}. 
Our work shows that augmenting NAT models the ability to handle the complex patterns of raw data (e.g., reordering patterns) with advanced training loss is a promising direction to accomplish the goal.

\iffalse
Non-autoregressive translation~\citet{guo2019non} has received increasing attention for its efficient decoding. However, such advantage also brings huge gap between NAT and AT models translation qualities.

To tackle this problem, several approaches try to build dependencies among the target tokens. The mainstream methods can mainly be divided into two different categories, namely iterative refinement and loss alignment. The first way builds target dependency via predict the target tokens round by round, e.g., CMLM~\cite{maskp} and CMLMC~\cite{huang2022improving}. However, multiple prediction rounds will bring time cost as shown in ~\citet{kasai2021deep}.
Loss alignment is mainly designed for fully NAT (i.e., only one decoding pass) to softens the penalty for word order errors. For example, ~\citet{axe} propose a new training loss which assigns loss based on the best possible monotonic alignment. Following this trend,~\citet{imputer} and ~\citet{Du:2021:ICML} propose different alignment methods for further relaxation. Our work extends \oaxe by relaxing at the phrase granularity, which shows improvement over both raw and distilled datasets.

\fi

\paragraph{Incorporating Ngrams into NMT}
Previous studies have incorporated the ngram phrases as an external signal to guide the generation in AT models~\cite{wang:2017:aaai,zhang2018prior,zhao2018phrase}. Concerning NAT models,~\newcite{guo2019non} enhance decoder inputs with ngram phrases,~\newcite{natcrf} use CRF to model bigram dependencies among target tokens to improve the decoding consistency.~\newcite{kong2020incorporatinglocal} use LSTM to generate ngram chunks, which are then merged via heuristic searching algorithm.
Closely related to our work,~\newcite{bowastarget} use bag of ngram phrases as additional training objective for AT models, and~\newcite{natbow} adapt this idea to NAT models. While~\newcite{natbow} require NAT models to fit all the possible orderings of ngrams, we compute the \noaxe loss based on the best ordering of ngrams.

\iffalse
Modeling ngram consistency is crucial for machine translation. One way to keep the ngram consistency is requiring model generates ngrams instead of tokens, e.g., ~\citet{natcrf} use CRF to generate bigram and ~\citet{kong2020incorporatinglocal} use LSTM to generate ngram chunk and then merge the chunk via heuristic searching algorithm. Another way is incorporating ngrams as supervision for the model.
Similar to our work, ~\citet{bowastarget} propose to use bag of ngram phrases as additional training objective for AT and ~\citet{natbow} adopt this idea into NAT filed.
However, our work is different from the previous two works. Because we only compute the best one reorderings of the whole ngram phrases while they require the model to fit all the possible reorderings.
\fi

\section{Conclusion}

In this work, we extend \oaxe by modeling ordering at the ngram phrase granularity, which can better ameliorate the effect of multimodality for NAT models. Benefiting from modeling translation at a larger granularity, the proposed \noaxe loss performs better at translating phrases and long sentences, and improves the fluency of generated translations. 
Extensive experiments on representative NAT benchmarks show that \noaxe consistently improves translation performance over \textsc{OaXE}, and is especially effective on raw data without distillation.

\section*{Acknowledgements}

We thank all anonymous reviewers for their insightful comments. And the authors gratefully acknowledge the support by the Lee Kuan Yew Fellowship awarded by Singapore Management University. 
\bibliography{ref}
\bibliographystyle{acl_natbib}

\iffalse
\clearpage
\appendix

\section{Appendix}
\subsection{Details of Syntactic Sequences}

\begin{table}[h]
    \centering
    \begin{tabular}{lrrrr}
    \toprule
    \multirow{2}{*}{\bf Model}   &   \multicolumn{4}{c}{\bf Syntactic Level}\\
    \cmidrule(lr){2-5}
        &   1   &   2   &   3 & 4\\
    \midrule
    \bf Gold Test Set & 1.0 & 1.5 & 2.2 & 3.0\\
    \midrule
    \multicolumn{5}{l}{\bf Raw Data}\\
    CMLM    & 1.0 &1.4 &1.9 &2.3\\
    \oaxe   & 1.0 &1.3 &1.9 &2.5\\
    \noaxe  & 1.0 &1.3 &1.9 &2.4\\
    \midrule
    \multicolumn{5}{l}{\bf Distilled Data}\\
    CMLM    & 1.0 &1.4 &2.0 &2.6\\
    \oaxe   & 1.0 &1.3 &1.9 &2.5\\
    \noaxe  & 1.0 &1.3 &1.9 &2.5\\
    \bottomrule
    \end{tabular}
    \caption{Averaged number of tokens that are covered by each tag at the syntactic sequence at different levels (from bottom to up). Statistics are calculated for the generated outputs on WMT14 En-De test set with multiple references.
    }
    \label{tab:average-tokens}
\end{table}

\fi

\end{document}